\documentclass[10pt,twocolumn,letterpaper]{article}

\usepackage[arxiv]{cvpr}             

\usepackage{float}
\usepackage{bm}
\usepackage{colortbl} 

\def\tgt{\text{tgt}}
\def\retr{\text{retr}}
\def\blend{\text{blend}}
\def\x{\bm{x}}

\def\p{\bm{p}}

\newcommand{\myparagraph}[1]{\smallskip\noindent\textbf{#1.}}

\definecolor{cvprblue}{rgb}{0.21,0.49,0.74}
\usepackage[pagebackref,breaklinks,colorlinks,allcolors=cvprblue]{hyperref}

\def\confName{}
\def\confYear{}

\title{ImageRAGTurbo: Towards One-step Text-to-Image Generation with Retrieval-Augmented Diffusion Models}

\author{Peijie Qiu, Hariharan Ramshankar, Arnau Ramisa, Ren\'e Vidal,\\ Amit Kumar K C, Vamsi Salaka, Rahul Bhagat\\
Amazon\\
 Core Search\\
{\tt\small \{peijieq, dsthari, aramisay, rvidal, amitkrkc, vsalaka, rbhagat\}@amazon.com}
}

\begin{document}
\maketitle
\begin{abstract}
    Diffusion models have emerged as the leading approach for text-to-image generation. However, their iterative sampling process, which gradually morphs random noise into coherent images, introduces significant latency that limits their applicability. While recent few-step diffusion models reduce the number of sampling steps to as few as one to four steps, they often compromise image quality and prompt alignment, especially in one-step generation. Additionally, these models require computationally expensive training procedures. To address these limitations, we propose ImageRAGTurbo, a novel approach to efficiently finetune few-step diffusion models via retrieval augmentation. Given a text prompt, we retrieve relevant text-image pairs from a database and use them to condition the generation process. We argue that such retrieved examples provide rich contextual information to the UNet denoiser that helps reduce the number of denoising steps without compromising image quality. Indeed, our initial investigations show that using the retrieved content to edit the denoiser's latent space ($\mathcal{H}$-space) without additional finetuning already improves prompt fidelity. To further improve the quality of the generated images, we augment the UNet denoiser with a trainable adapter in the $\mathcal{H}$-space, which efficiently blends the retrieved content with the target prompt using a cross-attention mechanism. Experimental results on fast text-to-image generation demonstrate that our approach produces high-fidelity images without compromising latency compared to existing methods.
\end{abstract}
    
\section{Introduction}
\label{sec:intro}

\begin{figure}
    \centering
    \includegraphics[width=1.0\linewidth]{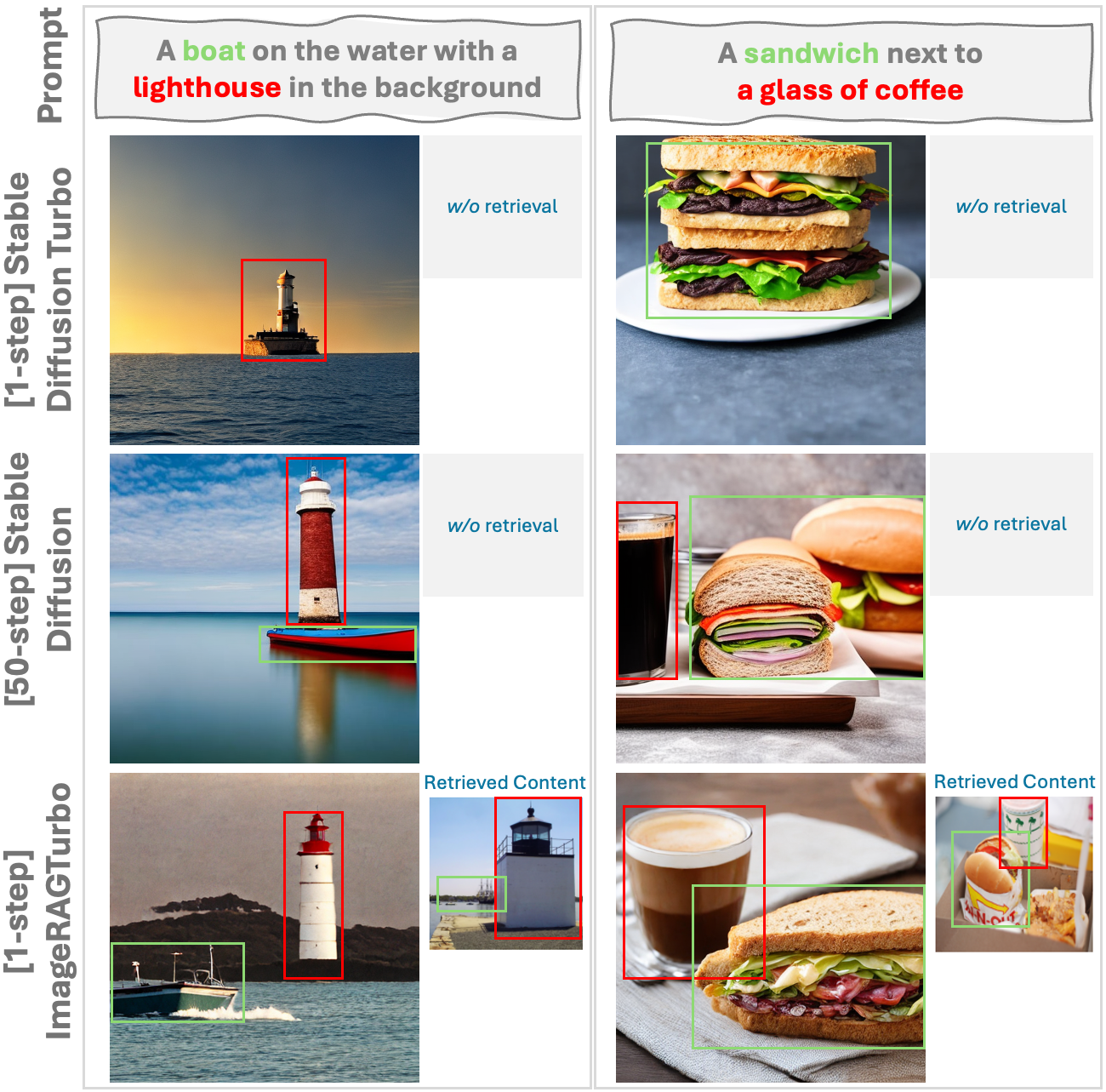}
    \caption{Illustrative examples of ImageRAGTurbo, where the visual concepts are highlighted by colored boxes. ImageRAGTurbo outperforms Stable Diffusion Turbo (adversarially distilled Stable Diffusion without retrieval) in generating accurate visual concepts. Even with a single step, ImageRAGTurbo performs comparably to Stable Diffusion with 50  steps. Please zoom in for better quality.}
    \label{fig:teaser}
\end{figure}

Diffusion models~\cite{ho2020denoisingdiffusionprobabilisticmodels,ramesh2022hierarchicaltextconditionalimagegeneration,rombach2022high,saharia2022photorealistic}  have demonstrated remarkable capabilities in generating high-quality images. However, their iterative sampling process  introduces a significant computational bottleneck that limits their practical applications. In particular, standard diffusion models typically require $25-50$ denoising steps to generate a single image, with each step involving an expensive network function evaluation of the denoiser. This sequential sampling process leads to high latency, limiting the use of these diffusion models in real-time and interactive applications. 

Classical approaches to improving the efficiency of diffusion models include (1) latent diffusion models that operate in a lower-resolution latent space to gain efficiency in the spatial dimension~\cite{rombach2022high}, and (2) fast ordinary differential equation (ODE) samplers that directly reduce the number of sampling steps~\cite{song2020denoising,lu2022dpm,lu2025dpm,zhao2023unipc}. However, despite remarkable progress in developing fast ODE samplers, they still require more than 10 steps to generate images with satisfactory quality.

Recent approaches to improving the efficiency of diffusion models include model quantization~\cite{li2023qa,li2023qb}, cached sampling~\cite{ma2024deepcache,zou2025accelerating}, and few-step distilled models~\cite{song2023consistency,luo2023latent,sauer2024adversarial,sauer2024fast}. Although the first two families of methods accelerate sampling, they still require a  large number of steps. In contrast, few-step models distill the full sampling trajectory of the teacher diffusion models to a short trajectory with just one to four steps. In the extreme one-step case, the distilled model learns to directly map noise to the target distribution. However, this poses the significant challenge of achieving high image and prompt fidelity\footnote{We use \textit{prompt fidelity} and \textit{text-to-image alignment} interchangeably.}~\cite{zhou2024simplefastdistillationdiffusion,sauer2024fast}. For example, given the prompt \textit{``A boat on the water with a lighthouse in the background.''}, the distilled few-step model may fail to generate the visual concept of \textit{``boat''} in the resulting images (see illustration in Fig.~\ref{fig:teaser}). Additionally, distilling these few-step models from teacher diffusion models typically requires substantial computational resources, making them less accessible for easy adoption.

One potential solution to improve prompt fidelity is to train preference models~\cite{kirstain2023pickapicopendatasetuser,wallace2023diffusionmodelalignmentusing}. However, this requires human annotated preference datasets. Another potential solution, drawing inspiration from retrieval-augmented generation (RAG) in large language models (LLMs)~\cite{lewis2020retrieval,ram2023context,edge2024local}, is to adapt RAG to enhance prompt fidelity in image generation~\cite{sheynin2022knndiffusionimagegenerationlargescale,blattmann2022semi,chen2022re,shalev2025imagerag,yuan2025finerag}. These models first query a database for relevant text-image pairs and then condition the diffusion models on the retrieved pairs to enhance prompt fidelity. Although prior retrieval-augmented image generation (RAIG) models have investigated  various conditioning strategies~\cite{sheynin2022knndiffusionimagegenerationlargescale,blattmann2022semi,chen2022re} and fine-grained retrieval content using LLMs~\cite{shalev2025imagerag,yuan2025finerag}, RAIG with few-step diffusion models has not yet been explored. In addition, most previous methods do not focus on improving efficiency in both training and inference~\cite{sheynin2022knndiffusionimagegenerationlargescale,blattmann2022semi,chen2022re,yuan2025finerag}. 

In this paper, we propose to use retrieval augmentation to improve the generation quality of few-step diffusion models while maintaining efficiency. The key intuition is that injecting semantically relevant retrieved information into the diffusion model can simplify the task of mapping random noise to target distributions. Preliminary results on using the retrieved content to directly edit the latent space of the diffusion model's denoiser ($\mathcal{H}$-space) without any finetuning already show improved image and prompt fidelity (see Fig.~\ref{fig:teaser}). However, this simple approach requires an expensive hyperparameter search at inference time to maximize performance, compromising the latency benefits of few-step image generation.
To mitigate this challenge, we propose to augment the denoiser's $\mathcal{H}$-space with a trainable adapter that blends the retrieved content with the target prompt using a cross-attention mechanism (Sec.~\ref{sec:method:finetuning}). 
Experiments show that the proposed ImageRAGTurbo method produces high-fidelity images without compromising efficiency.

Our main contributions can be summarized as follows:
\begin{itemize}
    \item We introduce a novel retrieval-augmented framework for few-step image generation with diffusion models, which significantly improves prompt fidelity and image quality while maintaining fast generation speeds.
     \item We develop an efficient finetuning approach using a lightweight adapter network in the $\mathcal{H}$-space, which reduces computational requirements compared to traditional few-step model training methods.
     \item We conduct extensive experiments demonstrating that our method outperforms existing few-step approaches in terms of both generation quality and prompt alignment, while maintaining comparable inference speed.
 \end{itemize}

The remainder of the paper is organized as follows: Section~\ref{sec:related_work} reviews relevant literature, Section~\ref{sec:method} details our method including our preliminary investigation in Section~\ref{sec:method:investigation} and the proposed $\mathcal{H}$-adapter finetuning in Section~\ref{sec:method:finetuning}. Section~\ref{sec:experiment} presents our experimental validation and results, and Section~\ref{sec:conclussion} concludes the paper.

\begin{figure*}[!t]
    \centering
    \includegraphics[width=1.0\linewidth]{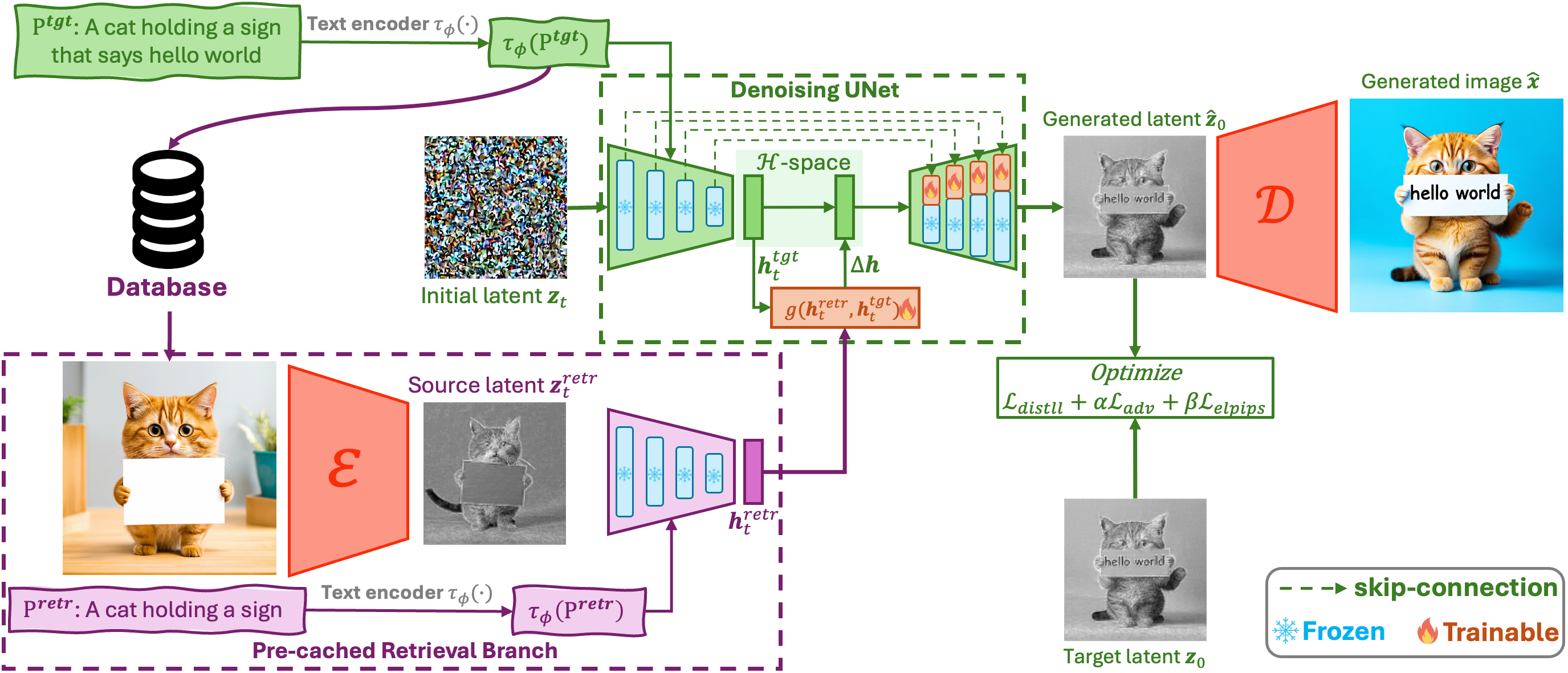}
    \caption{Overview of the proposed ImageRAGTurbo framework for efficiently finetuning the few-step diffusion models with retrieval-augmented generation. The framework involves two main branches: \textit{i)} a standard denoising branch (highlighted by green), and \textit{ii)} a retrieval branch (highlighted by purple). For a target prompt $\text{p}^{\bm{tgt}}$, it will first be converted to embeddings with a pretrained text encoder $\tau_\phi(\cdot)$. Then we query a database based on the target text embeddings to obtain the retrieved text-latent embeddings ($\tau_\phi(\text{p}^{\bm{retr}}), \bm{z}^{retr}_t$), which is then fed into the encoder of the denoiser to obtain the retrieved $\mathcal{H}$-space feature $\bm{h}_t^{retr}$. Finally, the $\mathcal{H}$-space feature $\bm{h}_t^{retr}$ in the retrieved branch is injected into the denoising branch by the proposed $\mathcal{H}$-space adapter to guide the generation.}
    \label{fig:framework}
\end{figure*}

\section{Related work}
\label{sec:related_work}

\myparagraph{Few-step Text-to-Image Generation}
Standard diffusion models~\cite{ho2020denoising} generate images via an iterative denoising process that aims to reverse the forward noising process step by step. This iterative reverse sampling process occurs sequentially and cannot be parallelized to speed it up. Although fast ODE samplers~\cite{song2020denoising,lu2022dpm,lu2025dpm,zhao2023unipc} can skip some steps by enabling a large sampling step size, these models cannot directly map noise to images like GANs~\cite{goodfellow2020generative,karras2020analyzing}. To address this challenge, consistency models~\cite{song2023consistency,luo2023latent} enforce the self-consistency of the denoising networks along the sampling trajectory such that they can map any point in an ODE trajectory to its starting point.
This enables one-step or few-step generation by directly mapping random noise to images, or through intermediate steps. However, despite being theoretically grounded, consistency models distilled from pretrained diffusion models, such as latent consistency models~\cite{luo2023latent}, typically result in low-quality samples when using few steps~\cite{sauer2024adversarial}. In contrast, adversarial diffusion distillation~\cite{sauer2024adversarial} and its variant in latent space~\cite{sauer2024fast} show strong empirical performance in few-step generation. There are also hybrid methods that combine consistency models and adversarial training~\cite{zhou2024simple,dao2025self}. 
However, all these methods share a common flaw: they often struggle to generate faithful images that align well with the target prompts.

\myparagraph{Retrieval-Augmented Image Generation}
RAG has been extensively explored in natural language processing~\cite{lewis2020retrieval,ram2023context,edge2024local}, but has received comparatively less attention in the context of image generation. 
An early example is the retrieval-augmented diffusion model (RDM)~\cite{blattmann2022semi}, 
which conditions the denoiser of a latent diffusion model~\cite{rombach2022high} on CLIP embeddings~\cite{radford2021learning} of image neighbors retrieved from an external database. 
Similarly, Re-Imagen~\cite{chen2022re} and its variant~\cite{hu2024instruct} extend the pixel-space training method of Imagen~\cite{saharia2022photorealistic} to condition the denoiser on the retrieved text-image pairs. In addition, RealRAG~\cite{lyu2025realrag} aims to enhance the retrieval branch of retrieval-augmented image generation by training a reflective retriever that selects retrieved images to complement the model’s missing knowledge. However, all these methods are not optimized for few-step image generation. Recent approaches like ImageRAG~\cite{shalev2025imagerag} and FineRAG~\cite{yuan2025finerag} leverage Multimodal LLMs (MLLMs) for dynamic image retrieval based on text prompts, and use these MLLMs to assess and progressively improve the generated images through iterative refinement. However, the repeated calls to MLLMs in these methods introduce substantial computational overhead, which defeats the primary goal of fast generation in this paper. 
\section{Method}
\label{sec:method}
In this work, we propose ImageRAGTurbo, a framework for boosting the performance of few-step generation models. Specifically, given a target prompt, $\p^{\tgt}$, we retrieve relevant text-to-image pairs $(\p^{\retr}, \x^{\retr})$ from a database. The retrieved information is then injected into the text-to-image diffusion models to guide the generation process. Motivated by the recent finding that the $\mathcal{H}$-space of the UNET denoiser already contains semantically meaningful representations, our investigation starts from a training-free direct $\mathcal{H}$-space injection (Sec.~\ref{sec:method:investigation}) and then extends to the proposed efficient $\mathcal{H}$-space adapter tuning (Sec.~\ref{sec:method:finetuning}). The overview of the proposed framework is shown in Fig.~\ref{fig:framework}.

\subsection{Preliminary}\label{sec:method:pre}

\myparagraph{Latent diffusion models}
In this work we focus on latent diffusion models (LDMs)~\cite{rombach2022high} and their popular successor stable diffusion models~\cite{podell2023sdxl,esser2024scaling}, rather than pixel-space alternatives like Imagen~\cite{saharia2022photorealistic}, due to their popularity and state-of-the-art performance for text-to-image generation. Specifically, LDMs have two stages: first, a pretrained VAE that transforms images $\bm{x}$ to their latent representations $\bm{z}$ via the encoder $\mathcal{E}: \bm{x} \mapsto \bm{z}$ and then back to the image space via the decoder $\mathcal{D}: \bm{z} \mapsto \bm{x}$; second, a diffusion model applied to the resulting latent space $\mathcal{Z}$. LDMs for text-to-image generation henceforth learn to progressively transform a standard Gaussian distribution $\bm{\epsilon} \sim \mathcal{N}(\mathbf{0}, \mathbf{I})$ along with a target prompt $\p^{\tgt}$ into the target latent $\bm{z}_0 = \mathcal{E}(\bm{x})$ through a denoiser $f_\theta$:
\begin{align}
\label{eq:parameterization}
    \hat{{\bm{z}}}_0 \!=\! f_\theta(\bm{z}_t, t, \tau_\phi(\p^{\tgt})), \bm{z}_t = \alpha_t \bm{z}_0 + \sigma_t \bm{\epsilon}, t\in \mathcal{U}(0, 1),\!
\end{align}
where $\alpha_t$ and $\sigma_t$ control the noise scheduling, and $\tau_\phi(\cdot)$ is a text encoder (\textit{e.g.,} a pretrained CLIP text encoder~\cite{radford2021learning}). 

Although there are different parameterizations of the denoiser~\cite{karras2022elucidating}, we opt for the one that predicts $\hat{\bm{z}}_0$ in Eq.~(\ref{eq:parameterization}), as it is widely used in few-step diffusion models~\cite{song2023consistency,luo2023latent,sauer2024adversarial,sauer2024fast}. The denoiser can be trained by optimizing the \textit{scoring matching} objective~\cite{vincent2011connection,song2020score}:
\begin{align}
    \min_\theta \mathbb{E}_{\bm{x} \sim p(\bm{x}), \bm{\epsilon}\sim \mathcal{N}(\bm{0}, \bm{I}), t\sim \mathcal{U}(0, 1)} \left[ \lambda(t)\| \hat{{\bm{z}}}_0 - \bm{z}_0 \|_2^2 \right],
\end{align}
where $p(\bm{x})$ is the data distribution, and $\lambda(t)$ is a weighting function. After training, we can generate the target latent $\hat{\bm{z}}_0$ by solving an SDE/ODE  using a variety of solvers~\cite{ho2020denoising,song2020denoising,lu2022dpm,lu2025dpm,zhao2023unipc}, starting from a random Gaussian noise $\bm{\epsilon} \sim \mathcal{N}(\mathbf{0}, \mathbf{I})$. The resulting latent $\hat{\bm{z}}_0$ is then projected back to the image space via the VAE decoder $\mathcal{D}$ to produce the final generated image $\hat{\bm{x}} = \mathcal{D}(\hat{\bm{z}}_0)$. 

\myparagraph{$\mathcal{H}$-space of the UNet denoiser} We define the $\mathcal{H}$-space of the denoiser~\cite{kwon2022diffusion}, which plays a crucial role in our method, as the space of feature maps $\bm{h}_t$ corresponding to the deepest layer of the UNet denoiser (see Fig.~\ref{fig:framework}). High-level concepts of a generated image, such as image class or object presence and attributes, are largely determined by $\mathcal{H}$-space features, which consequently impact the prompt fidelity~\cite{sclocchi2025phase}. In contrast, low-level features corresponding to earlier layers of the denoiser determine the details of a generated image.  

\subsection{\texorpdfstring{Retrieval-Augmented Direct $\mathcal{H}$-space Injection}{Investigation on H-space Manipulation with RAG}}
\label{sec:method:investigation}
Recent explorations have revealed that the denoising UNet already has a semantically meaningful $\mathcal{H}$-space~\cite{kwon2022diffusion,park2023understanding}. In light of this, we first investigate a simple training-free mechanism to directly inject the retrieved information into the $\mathcal{H}$-space of a  UNet denoiser to demonstrate the effectiveness of the retrieval augmented generation. Specifically, we first project the retrieved text-image pairs $(\p^{\retr}, \x^{\retr})$ to their corresponding embeddings $(\tau_\phi(\p^{\retr}), \bm{z}_0^{\retr})$, where $\bm{z}_0^{\retr} = \mathcal{E}(\bm{x}^{\retr})$ (see illustration in Fig.~\ref{fig:framework}). Subsequently, the retrieved embeddings $(\tau_\phi(\text{p}^{\retr}), \bm{z}_0^{\retr})$ are fed to the encoder of the UNet denoiser $f_\theta^{enc}$ to extract the $\mathcal{H}$-space features: $\bm{h}_t^{\retr} = f_\theta^{enc}(\tau_\phi(\text{p}^{\retr}), t, \bm{z}_0^{\retr})$. In practice, we set $t=0$, as the retrieved images are noise-free. Likewise, we can obtain the target $\mathcal{H}$-space features $\bm{h}_t^{\tgt} = f_\theta^{enc}(\tau_\phi(\text{p}^{\tgt}), t, \bm{z}_t)$.
We then enrich the target $\mathcal{H}$-space features by blending it with the retrieved $\mathcal{H}$-space features via spherical normalized interpolation:
\begin{equation}
    \bm{h}_t^{\blend} = \frac{\sin[(1- w) \Omega_t]}{\sin \Omega_t} \bm{h}^{\retr}_t + \frac{\sin[w \Omega_t]}{\sin \Omega_t} \bm{h}^{\tgt}_t,
\end{equation}
where $\Omega_t = \arccos ( \langle\bm{h}^{\tgt}_t, \bm{h}^{\retr}_t \rangle)$, and $w \in (0, 1)$ controls the strength of the blending. 
We employ spherical normalized interpolation rather than linear interpolation, as geodesic interpolation paths provide smoother semantic transitions~\cite{lobashev2025hessian}, effectively preventing the abrupt changes typically caused by phase transitions~\cite{sclocchi2025phase}.
The blended $\mathcal{H}$-space features along with the target prompt embeddings are then fed to the decoder of the UNet denoiser to produce the denoised latent $\hat{\bm{z}}_{t-1} = f_\theta^{dec}(\tau_\phi(\text{p}^{\tgt}), t, \bm{h}_t^{\retr})$.

\begin{figure}[!t]
    \centering   \includegraphics[width=1.0\linewidth]{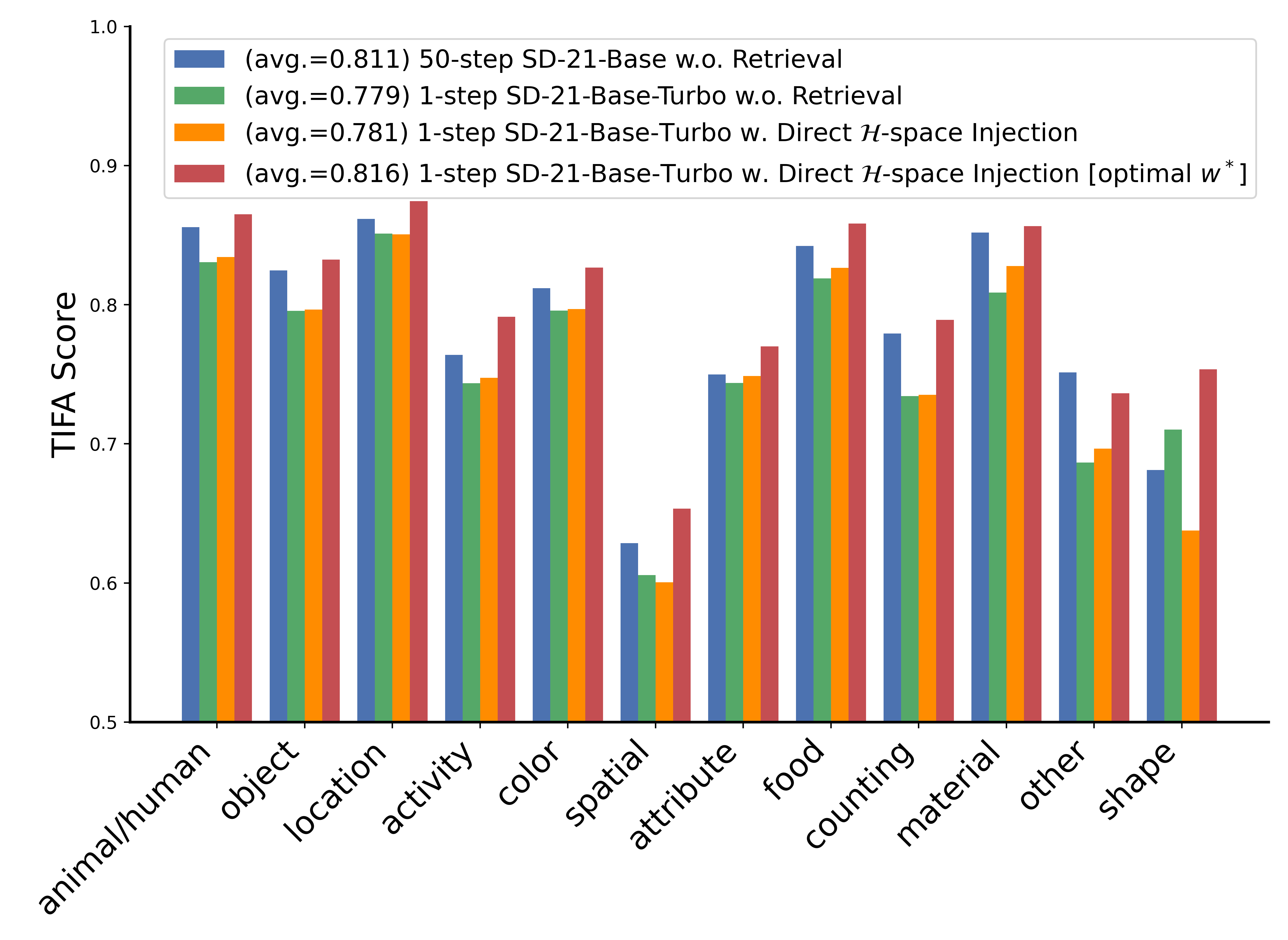}
    \caption{Performance of direct $\mathcal{H}$-space injection, shown as a histogram of TIFA scores across various categories.}
    \label{fig:tifa_dirc_injection}
\end{figure}

Our initial investigation has revealed that the direct $\mathcal{H}$-space injection without any further tuning yields a modest improvement in prompt fidelity (see Fig.~\ref{fig:tifa_dirc_injection}), increasing the TIFA score~\cite{hu2023tifa} from 0.779 to 0.781 with a fixed $w=0.8$ for all prompts. Observing that different prompts require different optimal blending strengths, we determine the optimal $w^*$ that maximizes the TIFA score for each prompt within a set of predefined values $\{0.1, 0.2, \ldots, 0.9 \}$ via brute force search/exhaustive search. This further improves the TIFA score to 0.816, even surpassing the 50-step stable diffusion model without retrieval augmentation. 
However, determining the optimal blending strength $w^*$ of each prompt is an ill-posed problem, as the optimal blending depends on latent factors such as retrieval relevance, prompt semantics, generation difficulty, \textit{etc}. A more principled solution to this problem is to edit the $\mathcal{H}$-space representations along multiple directions and compute the edit strength automatically. For example, recent work for LLMs~\cite{luo2024pace} constructs a large dictionary of latent editing directions (\textit{e.g.,} 40,000) and uses sparse coding to automatically select a small, relevant subset of directions along with their corresponding strengths. However, solving a sparse coding problem during inference can largely increase the inference time. Next, we propose an efficient method based on training an adapter that blends retrieved and target $\mathcal{H}$-space features.

\subsection{\texorpdfstring{Retrieval-Augmented Efficient $\mathcal{H}$-space Tuning}{RAG-based Efficient Adapter Tuning on H-space}}
\label{sec:method:finetuning}

Although directly injecting content into $\mathcal{H}$-space improves prompt fidelity, it requires exhaustive search for the optimal blending strength, making it impractical for real-world applications. This motivates us to automatically learn the correlations between the retrieved and target $\mathcal{H}$-space features. For this purpose, we introduce a trainable adapter $g_\varphi(\cdot, \cdot)$ to the $\mathcal{H}$-space of the UNet denoiser while freezing the other parts of the denoiser. The adapter $g_\varphi(\cdot, \cdot)$ leverages a cross-attention mechanism to automatically determine the correlations between the retrieved and target $\mathcal{H}$-space features:
\begin{align}
    \begin{split}
    \!\!
        & \quad \quad g_\varphi(\bm{h}^{\tgt}_t, \bm{h}^{\retr}_t) = \operatorname{softmax}\left(\frac{\bm{Q} \bm{K}^\top}{\sqrt{d_k}} \right) \bm{V}, \\
        & \bm{Q} = \bm{W}_Q \cdot \bm{h}^{\tgt}_t, \bm{K} = \bm{W}_K \cdot \bm{h}^{\retr}_t, \bm{V} = \bm{W}_V \cdot \bm{h}^{\retr}_t.
    \end{split}
    \!\!
\end{align}
The outputs from the adapter are then added to the raw target $\mathcal{H}$-space features as follows:
\begin{equation}
    \bm{h}_t^{\retr} = \bm{h}^{\tgt}_t + \lambda \underbrace{g_\varphi(\bm{h}^{\tgt}_t, \bm{h}^{\retr}_t)}_{\Delta \bm{h}_t},
\end{equation}
where $\lambda$ is a weighting parameter. We empirically find that the results are not sensitive to the choice of  $\lambda$, and set it to be the cosine similarity between the retrieved and target text CLIP embeddings. The intuition is that if the retrieved content has a high correlation to the target prompt, it should contribute more to the final generated content. 

\begin{figure}[!t]
    \centering
    \includegraphics[width=1.0\linewidth]{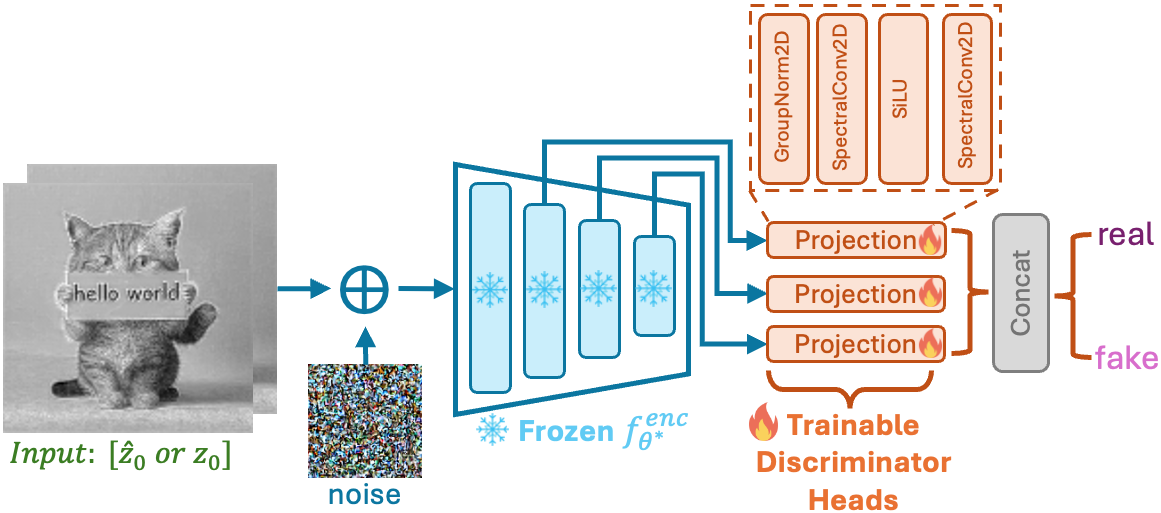}
    \caption{The architecture of the discriminator used for latent adversarial training, which adds noise to the samples $\hat{\bm{z}}_0$ from the student model and $\bm{z}_0$ from the teacher model, and differentiate them. 
    }
    \label{fig:discriminator}
\end{figure}

\subsection{Training Procedure}
Few-step diffusion models, trained through either self-consistency~\cite{song2023consistency,luo2023latent} or adversarial training~\cite{sauer2024adversarial,sauer2024fast}, operate on a subset of $n$ timesteps $T_{n} = \{t_1, t_2, \cdots, t_n\}$ drawn from the complete $N$ timesteps in the teacher model. For example, a 4-step student model uses $T_n = \{1.0, 0.75, 0.5, 0.25 \}$. The predicted $\hat{\bm{z}}_0$ at timestep $t_i, i\in \{1, \cdots, n\}$ of the proposed denoiser reads $\hat{\bm{z}}_0 = f_\theta(\bm{z}_{t_i}, t_i, \tau_\phi(\text{p}^{\tgt}), \bm{h}^{\retr}_0)$.
In this work, we focus on latent adversarial training~\cite{sauer2024fast}, as it empirically shows better performance than self-consistency training, especially in the one-step scenario. In addition, latent adversarial training, which operates on the latent $\mathcal{Z}$ space, is more efficient than its predecessor that performs adversarial training in the pixel space~\cite{sauer2024adversarial}. 
Without loss of generality, we denote the teacher and few-step student denoisers by $f_{\theta^*}$ and $f_{\theta}$, respectively,
both frozen in our training schema. 
The trainable module for the few-step denoiser is the $\mathcal{H}$-space adapter. To further help translate the adapted features into the final generation, we also finetune the decoder of the student denoiser using the parameter-efficient low-rank adaption tuning~\cite{hu2021loralowrankadaptationlarge}. 

In our latent adversarial training, the discriminator (denoted by $D$) consists of the frozen encoder of the teacher denoiser ($f_{\theta^*}^{enc}$) followed by a few trainable projection layers that project multi-stage encoder features to output embeddings (see Fig.~\ref{fig:discriminator}). Unlike standard adversarial training, latent adversarial training leverages the idea of DiffusionGAN~\cite{wang2022diffusiongan} that differentiates the noisy latent $\hat{\bm{z}}_s$ from $\bm{z}_s$, where $\hat{\bm{z}}_s=\alpha_s \hat{\bm{z}}_0 + \sigma_s \bm{\epsilon}$ and $\bm{z}_s=\alpha_s \bm{z}_0 + \sigma_s \bm{\epsilon}$, $s \sim \{t_1, \cdots, t_s, \cdots, t_N \}$. Following~\cite{sauer2021projectedgansconvergefaster,sauer2024adversarial,sauer2024fast}, the hinge loss is used as the adversarial objective function. The objective function for training the discriminator is given as:
\begin{align}
\begin{split}
\!\!
     \mathcal{L}^D_{\text{adv}} = \sum_k \mathbb{E}_{\bm{z}_0}\left[\max (0, 1 - D_k(\bm{z}_s, \tau_\phi(P^{\tgt}), t_s)) \right] + \\
     \mathbb{E}_{\hat{\bm{z}}_0} [ \max (0, 1 + D_k(\hat{\bm{z}}_s, \tau_\phi(P^{\tgt}), t_s)) ], 
\end{split}
\end{align}
where $D$ stands for discriminator including the frozen teacher encoder and trainable projection layers, and $k$ denotes the $k$-th stage features.
Similarly, the objective function for training the few-step denoiser (denoted by $G$) reads:
\begin{equation}
    \mathcal{L}^G_{\text{adv}} = - \sum_k \mathbb{E}_{\bm{z}_0} [D_k(\hat{\bm{z}}_s, \tau_\phi(\text{p}^{\tgt})].
\end{equation}
To stabilize the adversarial training procedure, we apply spectral normalization to the convolutional layers in the trainable projection heads of the discriminator (see Fig.~\ref{fig:discriminator}). Compared to the gradient penalty method used in~\cite{sauer2024adversarial,kang2024diffusion2gan}, spectral normalization does not require double backpropagation, and hence it saves computations. 

In addition to the adversarial training objective, we use two auxiliary objectives to stabilize training and improve image quality: \textit{i)} score distillation loss, and \textit{ii)} latent LPIPS loss~\cite{kang2024diffusion2gan}.  The score distillation loss is computed as the smoothed $L_1$ loss between $\hat{\bm{z}}_0$ and $\bm{z}_0$:
\begin{equation}
    \mathcal{L}_{\text{distill}}  = \begin{cases} 
    0.5* \| \hat{\bm{z}}_0 - \bm{z}_0 \|_2^2, \quad \text{if $|\hat{\bm{z}}_0 - \bm{z}_0| < 1$} \\
    | \hat{\bm{z}}_0 - \bm{z}_0  | - 0.5, \quad \text{otherwise}.
    \end{cases}
\end{equation}
Similar to the LPIPS loss, the latent LPIPS loss is computed in the latent space between $\hat{\bm{z}}_0$ and $\bm{z}_0$:
\begin{equation}
    \mathcal{L}_{\text{latentLPIPS}} = \mathbb{E}_{\mathcal{T}} \left[ \| F(\mathcal{T}(\hat{\bm{z}}_0)) - F(\mathcal{T}(\bm{z}_0) \|_2^2 \right],
\end{equation}
where $\mathcal{T}(\cdot)$ stands for a set of random differentiable augmentations, including general geometric transformations and cutoff, and $F(\cdot)$ the VGG embedding network~\cite{simonyan2015deepconvolutionalnetworkslargescale}.

The final training objective is the weighted sum of the aforementioned three objectives: 
\begin{equation}
    \mathcal{L} = \mathcal{L}_{\text{adv}} +  \alpha \mathcal{L}_{\text{distill}} + \beta \mathcal{L}_{\text{latentLPIPS}},
\end{equation}
where $\alpha$ and $\beta$ are weight-balance parameters. However, the extensive search of these parameters is expensive in practice. Following~\cite{sauer2024adversarial,kang2024diffusion2gan}, we empirically set $\alpha=2.5$ and $\beta=1.0$ in this paper.

\section{Experiments}
\label{sec:experiment}

\subsection{Experimental Setup}

\myparagraph{Finetuning dataset}
We finetune the proposed method using a mix of synthetic and real text-image pairs, as prior studies have shown that synthetic data can help improve prompt fidelity~\cite{sauer2024fast}. We generate synthetic data via the teacher model (\textit{i.e.,} Stable Diffusion v2-1-base~\cite{rombach2022high}) at a constant classifier-free guidance~\cite{ho2022classifierfreediffusionguidance} value of $7.5$, using around 3 million prompts from LAION-Aesthetic 6.25+ dataset~\cite{schuhmann2022laion5bopenlargescaledataset}. For the real data, we select 500K images from the LAION-Aesthetic 5.5+ dataset~\cite{schuhmann2022laion5bopenlargescaledataset} after filtering out those with resolutions smaller than $1024 \times 1024$. Both the real and synthetic images are subsequently resized to $512 \times 512$ and fed into the VAE encoder to obtain $64\times 64$ latent embeddings. 

\myparagraph{Evaluation benchmarks} We evaluate our proposed method on two widely used benchmarks: \textit{i)} MS-COCO~\cite{caesar2018cocostuffthingstuffclasses} and \textit{ii)} TIFA. Specifically, the MS-COCO benchmark uses $5,000$ text-image pairs from the validation set of the MS-COCO 2017 dataset, while the TIFA benchmark consists of $4,081$ text prompts that contain elements from 12 categories (\textit{e.g.,} object, spatial, activity, counting).

\myparagraph{Evaluation metrics} Our quantitative evaluation employs a set of metrics to assess different aspects of the generated images, including photorealism and faithfulness. For measuring the photorealism of the generated images, we consider the widely-adopted  FID score~\cite{heusel2017gans} on the COCO benchmark, which measures if the generated images follow the same distribution of the real images. In contrast, due to the absence of reference images, we use Aesthetics (AES) score to measure the visual appeal on the TIFA benchmark. To assess the faithfulness of the generated images to the prompts, we adapt the widely used CLIP score~\cite{hessel2021clipscore} on the COCO benchmark. We compute CLIP score using the \texttt{OpenCLIP-ViT-bigG-14} model. Although the CLIP score provides an overall assessment of text-to-image alignment, it may fail to provide an accurate evaluation of specific object attributes, quantities, and spatial relationships in the generated images. Therefore, we use the TIFA score~\cite{hu2023tifa} to measure if the generated images can truthfully reflect the target prompts by visual question answering.

\myparagraph{Baselines} To ensure a fair comparison, we compare the proposed method with several popular text-to-image LDM baselines with similar model sizes. In particular, we consider stable diffusion models~\cite{rombach2022high}, such as Stable Diffusion v1-5 and Stable Diffusion v2-1-base. In addition, we compare our model with few-step diffusion models, including the Latent Consistency Model~\cite{luo2023latent} and Stable Diffusion Turbo v2-1-base trained by latent adversarial distillation~\cite{sauer2024fast}. We also compare our method to RDM~\cite{blattmann2022semi} that integrates retrieval augmentation into stable diffusion models. To ensure a fair comparison, our method and all baselines have the same model size. For the distilled few-step models, the classifier-free guidance~\cite{ho2022classifierfreediffusionguidance} is disabled, as it will easily overshoot. For the other baseline models, we use a constant classifier-free guidance value of $7.5$.

\myparagraph{Implementation details} To obtain retrieved text-image pairs we apply the ScaNN search algorithm~\cite{guo2020acceleratinglargescaleinferenceanisotropic} in the text feature space of a
pretrained CLIP text encoder~\cite{radford2021learning}. Here, we reuse the same CLIP text features used by the stable diffusion model (\textit{i.e.,} \texttt{OpenCLIP-ViT-H-14}) without introducing an additional computational burden during inference. In addition, other retrieval methods, such as lexical prompt retriever (\textit{e.g.,} BM25 retriever) can be applied in different applications.
We use the 0.63M text-image pairs from the OpenImage database as the retrieval database. Our finetuning is performed on 64$\times$ NVIDIA L40S GPUs with a total batch size of $2,048$. We finetune the 1-step stable diffusion turbo model for 20K iterations using the AdamW optimizer with a constant learning rate of $1\times 10^{-5}$ for both few-step student model and the discriminator.

\begin{table}
\centering
\caption{Quantitative comparison of text-to-image generation models on the MS-COCO benchmark measured by the FID and CLIP scores, as well as the number of function evaluations (NFE) in the denoising process.}
\resizebox{1.\columnwidth}{!}{
\begin{tabular}{lccc}
\hline
Models & NFE & FID$\downarrow$ & CLIP$\uparrow$ \\
\hline
\multicolumn{4}{l}{\textit{w/o} Retrieval} \\
\hline
Stable Diffusion v1-5 & 50 & 24.38 & 0.319 \\
Stable Diffusion v2-1-base & 50 &  25.33 & 0.330 \\
Latent Consistency Model & 4 & 36.52 & 0.307 \\
Stable Diffusion Turbo v2-1-base & 1 & 26.04 & 0.319 \\

\hline
\multicolumn{4}{l}{\textit{w/} Retrieval} \\
\hline
RDM & 50 & 27.60 & 0.293 \\
\rowcolor{blue!8}
ImageRAGTurbo (Ours) & 1 & 25.59 & 0.323 \\
\hline
\end{tabular}
}
\label{tab:coco}
\end{table}

\begin{table}
\centering
\caption{Quantitative comparison of text-to-image generation models on the TIFA benchmark as measured by the TIFA and AES scores, and the number of function evaluations (NFE) in denoising.\!}
\resizebox{1.\columnwidth}{!}{
\begin{tabular}{lccc}
\hline
Models & NFE & AES$\uparrow$ & TIFA$\uparrow$ \\
\hline
\multicolumn{4}{l}{\textit{w/o} Retrieval} \\
\hline
Stable Diffusion v1-5 & 50 & 5.79 & 0.768 \\
Stable Diffusion v2-1-base & 50 & 6.04 & 0.811 \\
Latent Consistency Model & 4 & 5.80 & 0.764 \\
Stable Diffusion Turbo v2-1-base & 1 & 5.85 & 0.779 \\

\hline
\multicolumn{4}{l}{\textit{w/} Retrieval} \\
\hline
RDM & 50 & 5.40 &  0.725 \\
\rowcolor{blue!8}
ImageRAGTurbo (Ours) & 1 & 5.88 & 0.801 \\
\hline
\end{tabular}
}
\label{tab:tifa}
\end{table}

\begin{figure*}[!t]
  \centering
   \resizebox{400pt}{180pt}{
    \includegraphics[width=1.0\linewidth]{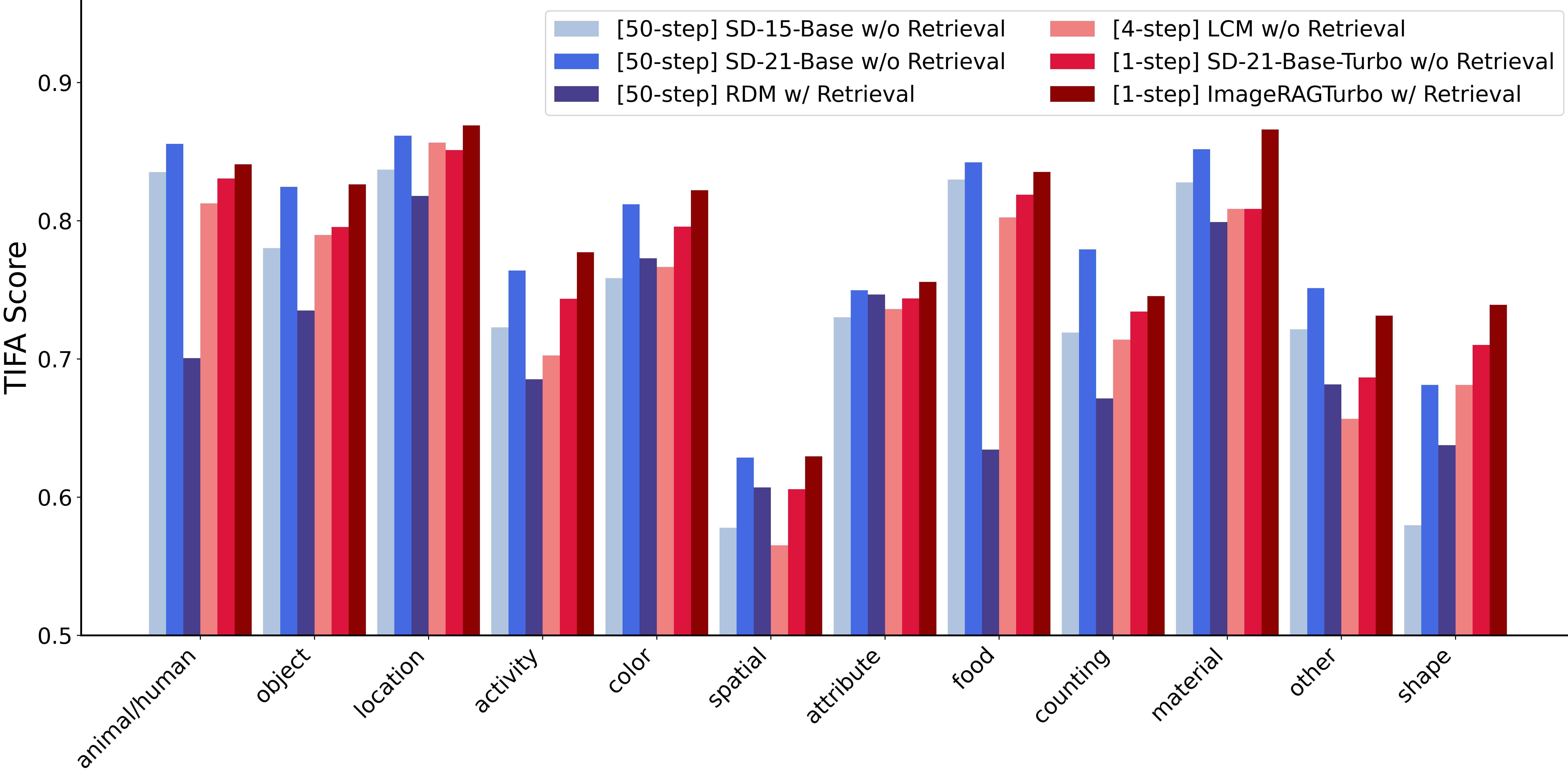}
    }
  \caption{Detailed histogram of TIFA scores across various categories. Despite still lagging behind 50-step Stable Diffusion v2-1-base model, our 1-step ImageRAGTurbo achieves comparable or even slightly higher TIFA score in certain categories such as \textit{object, activity, and material}.}
  \label{fig:tifa_hist}
\end{figure*}

\subsection{Results on the MS-COCO Benchmark}
Our evaluation on the MS-COCO benchmark demonstrates ImageRAGTurbo's superior performance (see Table~\ref{tab:coco}). Compared to the baseline Stable Diffusion Turbo v2-1-base without retrieval augmentation, our approach achieves better text-image alignment with a CLIP score that is 1.37\% higher, and improved image quality as indicated by a lower FID score.
The single-step ImageRAGTurbo significantly outperforms the 4-step latent consistency model in both prompt fidelity and image quality metrics. Furthermore, our 1-step model achieves comparable results to the 50-step teacher model (\textit{i.e.,} Stable Diffusion v2-1-base), showing a drop of 1.03\% and 2.1\% in FID and CLIP scores, respectively.
These improvements can be attributed to ImageRAGTurbo's enhanced ability to retrieve and leverage fine-grained, high-quality reference images for more precise image synthesis. In comparison to other retrieval-augmented approaches, ImageRAGTurbo shows substantial improvement over RDM with a CLIP score that is 10\% higher. 

\subsection{Results on the TIFA Benchmark}
As shown in Table~\ref{tab:tifa}, ImageRAGTurbo also demonstrates superior performance on the TIFA benchmark. Compared to the baseline Stable Diffusion Turbo v2-1-base without retrieval augmentation, our approach achieves better text-image fidelity with a TIFA score that is 2.2\% higher, and a slightly better aesthetic quality as indicated by the aesthetic score (5.88 vs 5.83).
The single-step ImageRAGTurbo significantly outperforms the 4-step latent consistency model, showing a 3.7\% improvement in TIFA score. Although our 1-step model shows an overall drop of 1.2\% in TIFA score compared to 50-step Stable Diffusion v2-1-base model, it achieves a comparable or even slightly higher TIFA score in certain categories such as \textit{object, location, activity, and material} (see Fig.~\ref{fig:tifa_hist}). 
Compared to other retrieval-augmented approaches, ImageRAGTurbo substantially outperforms RDM with a TIFA score 7.6\% higher, and a better aesthetic quality (5.88 vs 5.40), while requiring only a single inference step instead of 50 steps. 

\begin{figure}[!t]
    \centering
    \includegraphics[width=1.0\linewidth]{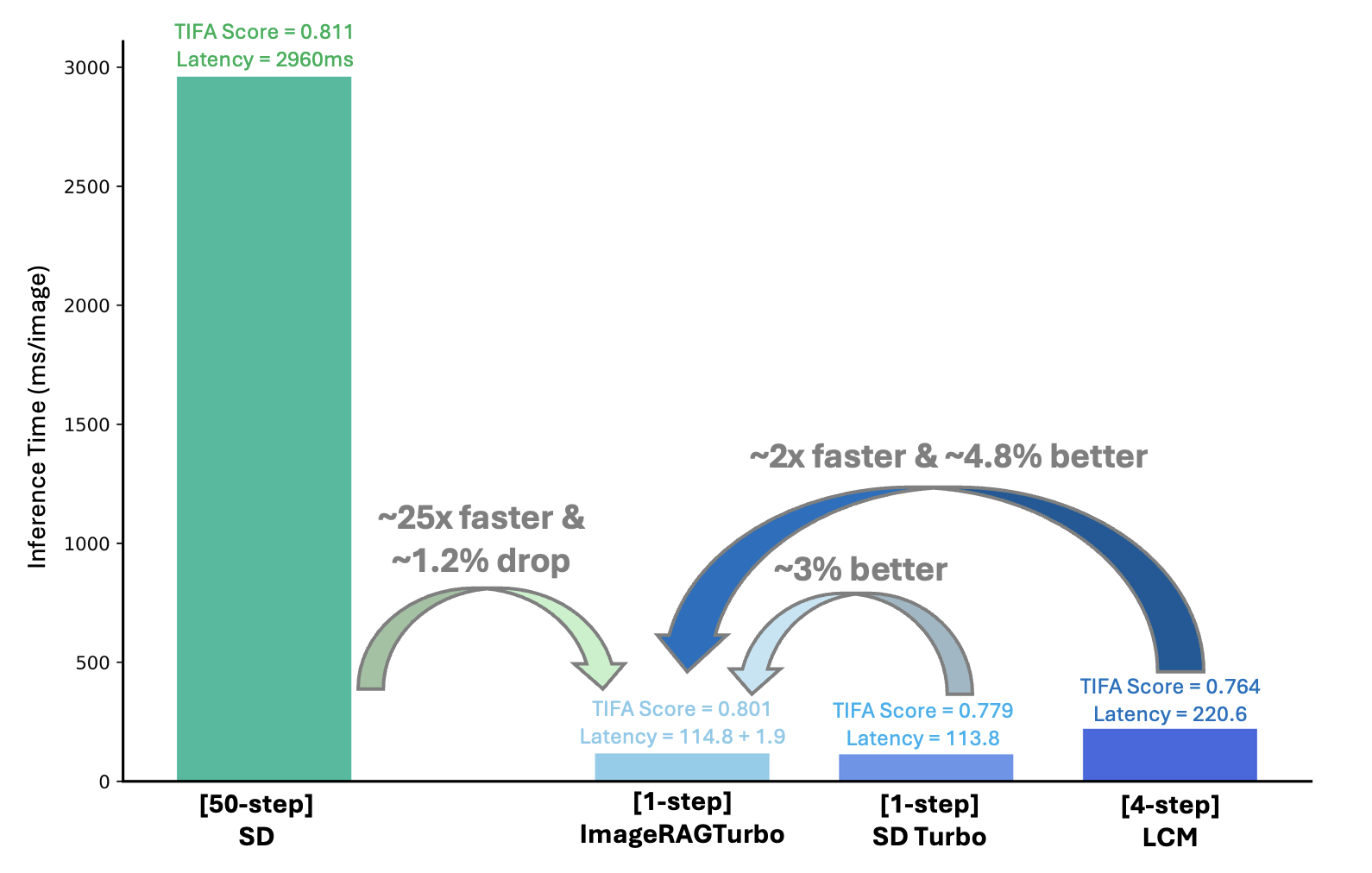}
    \caption{Comparison of inference time (ms$/$image). We report the average inference time calculated over the same set of 100 prompts at $512 \times 512$ resolution on a single NVIDIA L40S GPU.}
    \label{fig:effiency}
\end{figure}

\begin{figure*}[!t]
  \centering
  \resizebox{500pt}{340pt}{
    \includegraphics[width=1.0\linewidth]{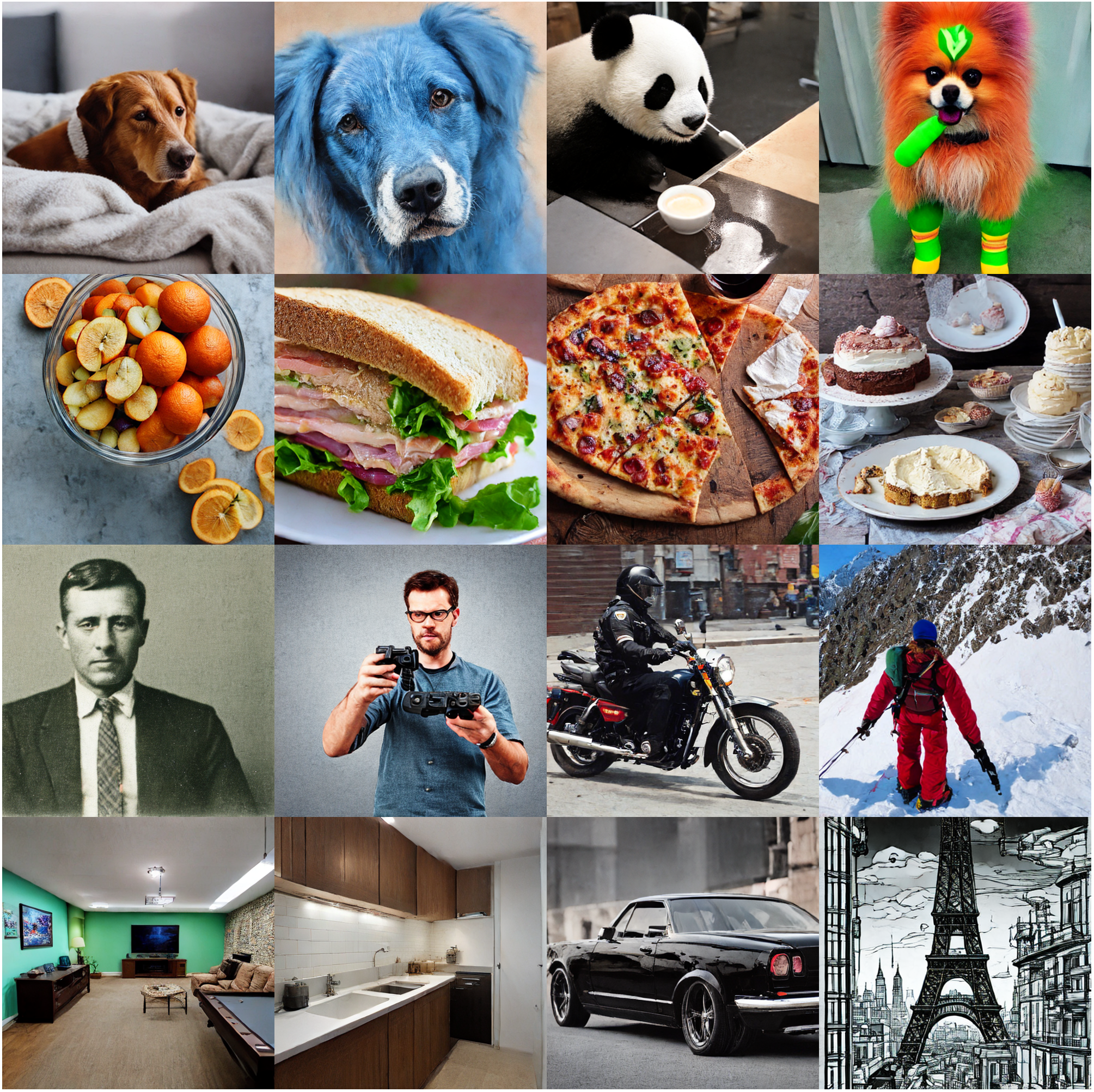}
    \includegraphics[width=1.0\linewidth]{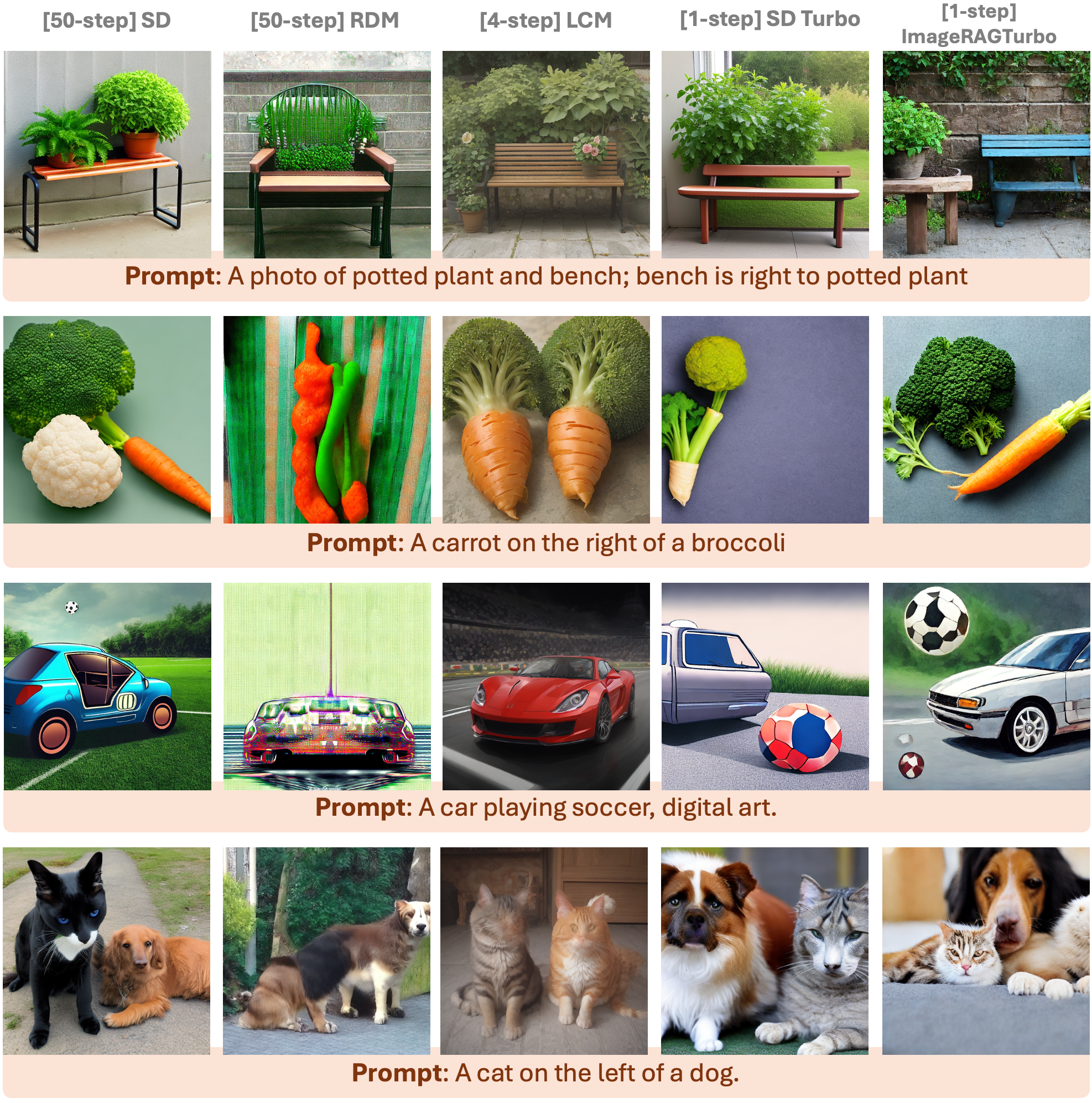}
  }
  \caption{Qualitative results of ImageRAGTurbo. ImageRAGTurbo can generate high-quality images in a single step (\textbf{Left}) and achieves better text-to-image alignment compared to other competing few-step methods (\textbf{Right}).}
  \label{fig:qualitative}
\end{figure*}

\subsection{Efficiency Analysis}
Fig.~\ref{fig:effiency} shows the computational efficiency of our method compared to baselines. The total inference time of our method (116.7ms) is on par with that of the Stable Diffusion (SD) Turbo (113.8ms). The inference time of our method can be broken down into two components: retrieval (1.9ms$/$image) and denoising (114.8ms$/$image). The 1 ms increase in denoising latency compared to SD Turbo is attributable to the $\mathcal{H}$-space adapter, with 36M additional parameters, representing only 4\% of the total model parameters. Our method introduces only a minimal overhead of 2.5\% in latency compared to the SD Turbo, due to the added retrieval mechanism and adapter, while delivering a significant 3\% improvement in TIFA score. Additionally, the proposed method is approximately 25$\times$ faster than the SD teacher model, with only a 1.2\% drop in TIFA score.
 Compared to the latent consistency model, our method demonstrates a 3.7\% higher TIFA score with around 47\% more latency (116.7ms vs 220.6ms).

\subsection{Qualitative Analysis}
To highlight the effectiveness of our proposed model and its potential applications, we present qualitative examples in Fig.~\ref{fig:qualitative}. Specifically, our approach can generate high-quality images in a single step (see Fig.~\ref{fig:qualitative}, \textbf{Left}). In addition,
our approach outperforms other few-step baselines in text-to-image alignment (see Fig.~\ref{fig:qualitative}, \textbf{Right}). In particular, our RAG-based proposed method can generate accurate objects (demonstrated in \textit{examples/rows 2 and 3}) and proper spatial relationship (shown in \textit{examples/rows 1 and 4}).
\section{Conclusion and Future Work}
\label{sec:conclussion}

We have presented ImageRAGTurbo, a novel retrieval-augmented framework designed to enhance few-step diffusion models through efficient finetuning. Our approach tackles a key challenge in few-step diffusion models (insufficient text-to-image alignment) by effectively leveraging relevant information from a retrieval database. Through this efficient retrieval-augmented mechanism, ImageRAGTurbo can improve the model's ability to generate images that faithfully reflect the input prompts without compromising the inference latency. Experimental results show that our method significantly improves text-to-image alignment without compromising image quality. While our framework paves the way for retrieval-augmented few-step diffusion models, there remains significant potential for further advancement. Our current implementation relies on CLIP-based retrieval; however, the framework could benefit from more fine-grained retrieval. In particular, exploring compositional retrieval approaches represents a promising future direction.
{
    \small
    \bibliographystyle{ieeenat_fullname}
    \bibliography{main}
}

\end{document}